\documentclass{article}

\usepackage[preprint]{neurips_2023}

\usepackage[utf8]{inputenc} 
\usepackage[T1]{fontenc}    
\usepackage[breaklinks,colorlinks,bookmarks=false]{hyperref}
\usepackage{url}            
\usepackage{booktabs}       
\usepackage{amsfonts}       
\usepackage{nicefrac}       
\usepackage{microtype}      
\usepackage{xcolor}         
\usepackage{amsmath}
\usepackage{wrapfig}

\usepackage[capitalize]{cleveref}
\crefname{section}{Sec.}{Secs.}
\Crefname{section}{Section}{Sections}
\Crefname{table}{Table}{Tables}
\crefname{table}{Tab.}{Tabs.}

\usepackage{bm}
\usepackage{amssymb} 
\usepackage{amsthm}
\usepackage{algorithm}
\usepackage{algorithmic}
\usepackage{multirow}
\usepackage{graphicx}
\usepackage{subcaption}
\usepackage{xcolor}         
\usepackage{colortbl}

\newtheorem{proposition}{Proposition}

\usepackage{pifont}
\newcommand{\cmark}{\ding{51}}%
{ \bgroup
  \addtolength\abovedisplayshortskip{#1}
  \addtolength\abovedisplayskip{#1}
  \addtolength\belowdisplayshortskip{#1}
  \addtolength\belowdisplayskip{#1}}
{\egroup\ignorespacesafterend}

\newcommand{\name}{BAPE}

\title{
    BAPE: Learning an Explicit Bayes Classifier \\
    for Long-tailed Visual Recognition
}

\author{%
  Chaoqun Du \quad Yulin Wang \quad Shiji Song \quad Gao Huang \\
  Department of Automation\\
  Tsinghua University\\
  Beijing, China \\
  \texttt{dcq20@mails.tsinghua.edu.cn} \quad \texttt{gaohuang@tsinghua.edu.cn}
}

\begin{document}

\maketitle

\begin{abstract}

Bayesian decision theory advocates the Bayes classifier as the optimal approach for minimizing the risk in machine learning problems. Current deep learning algorithms usually solve for the optimal classifier by \emph{implicitly} estimating the posterior probabilities, \emph{e.g.}, by minimizing the Softmax cross-entropy loss. This simple methodology has been proven effective for meticulously balanced academic benchmark datasets. However, it is not applicable to the long-tailed data distributions in the real world, where it leads to the gradient imbalance issue and fails to ensure the Bayes optimal decision rule.
To address these challenges, this paper presents a novel approach (BAPE) that provides a more precise theoretical estimation of the data distributions by \emph{explicitly} modeling the parameters of the posterior probabilities and solving them with point estimation. Consequently, our method directly learns the Bayes classifier without gradient descent based on Bayes' theorem, simultaneously alleviating the gradient imbalance and ensuring the Bayes optimal decision rule.
Furthermore, we propose a straightforward yet effective \emph{distribution adjustment} technique. This method enables the Bayes classifier trained from the long-tailed training set to effectively adapt to the test data distribution with an arbitrary imbalance factor, thereby enhancing performance without incurring additional computational costs.
In addition, we demonstrate the gains of our method are orthogonal to existing learning approaches for long-tailed scenarios, as they are mostly designed under the principle of \emph{implicitly} estimating the posterior probabilities.
Extensive empirical evaluations on CIFAR-10-LT, CIFAR-100-LT, ImageNet-LT, and iNaturalist demonstrate that our method significantly improves the generalization performance of popular deep networks, despite its simplicity.

\end{abstract}

\section{Introduction}
\label{sec:intro}

As grounded in Bayesian decision theory, the Bayes classifier is usually identified as the optimal classifier that minimizes the risk in machine learning tasks \cite{vapnik1991principles,vapnik1999nature,murphy2012machine}.
However, the complex and often intractable nature of real data distributions presents significant challenges to the direct computation of the posterior distribution, and hence hinders attaining Bayes-optimal decision-making.
In most cases, popular methodologies are designed to \emph{implicitly} estimate posterior probabilities based on training data. This typically involves the optimization of a Softmax classifier through the application of cross-entropy loss and gradient descent \cite{Goodfellow-et-al-2016,Krizhevsky2017,He2016,Huang2017,dosovitskiy2021an,guo2023faceclip,guo2024smooth,guo2023zero,guo2022assessing}. This approach has consistently demonstrated its efficacy across a broad range of benchmark datasets \cite{deng2012mnist, krizhevsky2009learning, Russakovsky2015}.



However, real-world application scenarios are generally not as ideal as these academic benchmarks \cite{du2024unitta,du2024switta,guo2025everything,du2024probabilistic,du2024simpro,huang2022high}.
As a notable difference, real data often follows a long-tailed distribution, characterized by a significant drop in the number of samples per class from the head (high-frequency classes) to the tail (low-frequency classes) \cite{VanHorn2018}.
Given such an imbalance, the standard Softmax cross-entropy algorithm suffers from two major issues. 
Through the lens of optimization, it leads to a \emph{minority collapse} phenomenon \cite{Fang2021} caused by gradient imbalance, \emph{i.e.}, classifiers for minority classes tend to become closer to each other due to the progressively suppressed gradients as the level of imbalance escalates.
From the perspective of Bayesian decision-making, the Bayesian optimal decision rule is usually not ensured \cite{Menon2021}.
To alleviate these two problems, some methods have been proposed, such as re-sampling \cite{Kubat1997,Wallace2011,Chawla2002,Kubat1997,buda2018systematic,byrd2019effect,mahajan2018exploring,Kang2020}, re-weighting \cite{Huang2016,Menon2013,Cui2019,lin2017focal,Wang2021b,wu2020distribution,Tan2020}, re-margining \cite{Cao2019,khan2019striking,cao2020domain} and logit adjustment \cite{hong2021disentangling, Menon2021, provost2000machine, wang2022towards, Ren2020}.
However, importantly, they continue to adopt the paradigm that \emph{implicitly} estimates the posterior distribution. Such implicit estimation is developed and mostly effective for balanced training data. 
We argue that this design may result in sub-optimal algorithms in the context of realistic long-tailed training data.

In this paper, we seek to explore whether a better machine learning algorithm can be acquired under the principle of \emph{explicitly} modeling the Bayesian decision process. We achieve this by proposing a \name~approach (BAyes classifier by Point Estimation). In specific, \name~offers a more precise theoretical estimation of data distribution by \emph{explicitly} modeling the data distribution and employing point estimation for distribution parameters. As a consequence, it directly learns the Bayes classifier through point estimation without relying on gradient descent, effectively mitigating the issue of gradient imbalance. Furthermore, the Bayes classifier is learned explicitly based on Bayes' theorem, adhering to the optimal decision rule.

Our primary focus lies in explicitly modeling the data distribution and estimating its parameters. However, translating this idea into practice is not straightforward, as the methodologies for modeling real data distributions often involve complexities, such as the necessity of training deep generative models \cite{Goodfellow2020,SohlDickstein2015}.
We propose a more elegant and manageable solution to this challenge. Instead of modeling the data distribution in its original form, we propose to do so in the feature space, which is generally simpler and more tractable.
Specifically, inspired by the Simplex-Encoded-Labels Interpolation (SELI) \cite{Thrampoulidis2022} that the features tend to collapse towards their corresponding class means in imbalanced learning, we adopt the von Mises-Fisher (vMF) distribution on the unit sphere to model the feature distribution.
Building on this vMF distribution assumption, we can estimate the parameters using point estimation method. Crucially, this only requires the first sample moment, which can be efficiently computed across various batches during the training process. Thus, we eliminate the need for gradient descent, making our approach computationally efficient and straightforward.

Moreover, a crucial aspect we consider is the disparity in feature distributions between the training and testing set, a phenomenon often arising due to the limited size of available data for estimation. Such discrepancy implies that a Bayes classifier trained on the training set might not deliver optimal performance when applied to the test set.
To counter this, leveraging our explicit estimation of data distribution, we can modify the parameters to better align with the test set which exhibits an arbitrary imbalance factor.
Notably, our approach incurs no additional computational costs, making it an efficient solution. Empirical results demonstrate that it considerably enhances the performance.

The primary contributions of this study are outlined as follows:
1) we propose a novel method for explicitly learning a Bayes classifier using point estimation, which ensures the Bayesian optimal decision rule and alleviates the problem of gradient imbalance;
2) we introduce a straightforward yet effective method to adjust the distribution, which boosts performance without incurring any additional costs;
3) empirical evaluations on image classification tasks with CIFAR-10/100-LT, iNaturalist 2018, and ImageNet-LT demonstrate that the proposed \name~algorithm consistently improves the generalization performance of existing long-tailed recognition methods.
\section{Related Works}

\textbf{Re-sampling.} The re-sampling methods aim to rectify the uneven distribution of training data by either downsampling the high-frequency classes \cite{Kubat1997,Wallace2011} or upsampling the low-frequency classes \cite{Chawla2002,buda2018systematic,byrd2019effect}, thereby facilitating the acquisition of knowledge of the tail classes.
Square-root sampling \cite{mahajan2018exploring} is a modified version of class-balanced sampling, in which the sampling probability for each class is determined by the square root of the sample size within that specific class.
Progressively-balanced sampling \cite{Kang2020} gradually transitions between random sampling and class-balanced sampling.
Empirical findings from Decoupling \cite{Kang2020} demonstrate that both square-root sampling and progressively-balanced sampling are superior strategies for training standard models in long-tailed recognition.

\textbf{Re-weighting.}
In order to mitigate the impact of class imbalance, re-weighting techniques strive to adjust the training loss values associated with various classes by multiplying distinct weight factors \cite{Huang2016, Cui2019, lin2017focal, Wang2021b, wu2020distribution}.
Following this methodology, Class-balanced loss (CB) \cite{Cui2019} introduces an effective number term to approximate the expected sample count for distinct classes and 
then incorporates a re-weighting term that balances classes by inversely scaling it with the effective number.
Some recent studies \cite{Tan2020, Wang2021b, wu2020distribution} also seek to address the negative gradient over-suppression issue of tail classes by re-weighting.
Equalization loss \cite{Tan2020} simply reduces the impact of tail-class samples when they act as negative labels for head-class samples. 

\textbf{Re-margining.}
In tackling class imbalance, re-margining techniques attempt to modify losses by subtracting distinct margin factors for different classes.
Following this idea, LDAM \cite{Cao2019} incorporates class-specific margin factors determined by the training label frequencies, encouraging larger margins for tail classes.
Recent studies further explored adaptive re-margining methods.
Uncertainty-based margin learning (UML) \cite{khan2019striking} utilizes estimated class-level uncertainty to adjust loss margins. 
A subsequent work introduces a frequency indicator based on the inter-class feature compactness \cite{cao2020domain}.

\textbf{Logit Adjustment.}
Logit adjustment techniques \cite{hong2021disentangling, Menon2021, provost2000machine, wang2022towards, Ren2020} aim to address the class imbalance by modifying the prediction logits of a class-biased model.
In a recent study \cite{Menon2021}, a comprehensive analysis of logit adjustment in the context of long-tailed recognition was conducted, proposing a logit adjustment (LA) method from the Bayesian perspective, which ensures the Bayesian optimal decision rule.
Most recently, a vMF classifier \cite{wang2022towards} is introduced, which performs adjustments via inter-class overlap coefficients. 
While this approach shares similarities with our method in terms of utilizing the vMF distribution and employing a post-hoc adjustment technique,  it involves complex adjustments to enhance model performance, albeit at the expense of simplicity.
This method fundamentally differs from our approach in that it still relies on gradient descent for learning distribution parameters and is susceptible to gradient imbalance.

\section{Method}
\label{sec:method}

In this section, we first introduce the theoretical motivation behind our approach based on Bayesian optimal decision-making. Next, we introduce the assumption and its rationale concerning the modeling of the distribution, which allows us to efficiently estimate the classifier's parameters by point estimation.
Subsequently, an algorithm is presented for efficiently estimating parameters using maximum a posteriori probability estimation during the training process.
Moreover, we propose a distribution adjustment method for adapting the Bayes classifier to the testing process.

\subsection{Theoretical Motivation}
\textbf{Preliminaries.} We start by presenting the problem setting, laying the basis for introducing our method.
Given the training set $\mathcal{D} = \{{(\bm{x}_i, y_i)}_{i=1}^N\}$, the model is trained to map the images from the space $\mathcal{X}$ into the classes from the space $\mathcal{Y} = \{1,2,\dots,K\}$.
Typically, the mapping function $\varphi$ is modeled as a neural network, which consists of a backbone feature extractor $F\colon \mathcal{X}\rightarrow \mathcal{Z}$ and a linear classifier $G\colon \mathcal{Z}\rightarrow \mathcal{Y} \colon z\mapsto \arg\max (\bm{W}^T\bm{z} + \bm{b})$.
The standard Softmax cross-entropy loss for a sample $\{\bm{x},y\}$ in training set can be expressed as:
\begin{equation}
    \mathcal{L}_\text{Softmax}(\bm{x}) = -\log p(y|\bm{x}), \quad p(y|\bm{x}) = \frac{\exp (\bm{w_{y}}^T\bm{z} + b_y)}{\sum_{y'} \exp(\bm{w_{y'}}^T\bm{z}+b_{y'})},
    \label{eq:softmax}
\end{equation}
where $\bm{w_y}$ and $b_y$ are the weight and bias of the linear classifier for class $y$, respectively.
It can be observed from Eq.~(\ref{eq:softmax}) that the model \emph{implicitly} estimates the posterior probability of the class.

\textbf{The Bayes Classifier} is the optimal classifier that minimizes the probability of misclassification, which is defined based on the Bayes' theorem:
\begin{equation}
    p(y|\bm{x})=\frac{p(y)p(\bm{x}|y)}{\sum_{y'}p(y')p(\bm{x}|y')}.
     \label{eq:BC}
\end{equation}
However, capturing the true distribution of real-world data proves challenging due to its inherent complexity.
As a result, most existing approaches resort to approximating the Bayes classifier by the model's output, as illustrated in \cref{eq:softmax}.
Nevertheless, in long-tail recognition tasks, the test sets are typically  balanced, resulting in a disparity in the prior distribution of $p(y)$ between the training and test sets.
The long-tail recognition approaches for balancing the gradients overlook this difference, thereby failing to ensure the model learns the Bayes optimal classifier.
To tackle this issue, Logit Adjustment \cite{Menon2021} introduces a prior distribution over the class labels:
\begin{equation}
    \mathcal{L}_{LA}(\bm{x}) = -\log p(y|\bm{x}), \quad p(y|\bm{x}) = \frac{\pi_y \exp (\bm{w_{y}}^T\bm{z} + b_y)}{\sum_{y'} \pi_{y'} \exp(\bm{w_{y'}}^T\bm{z}+b_{y'})},
    \label{eq:LA}
\end{equation}
where $\pi_y$ is the class frequency in the training or test set.
Nonetheless, these methods all implicitly estimate the posterior probability of the classes using gradient descent, without leveraging information regarding the data distribution.

\subsection{Distribution Assumption}
\label{sec:assumption}

In this section, we first introduce the distribution assumption underlying our method and explain its rationality.
Based on this assumption, we present the specific form of the Bayes classifier.

As mentioned earlier, the complexity of real-world data makes it difficult to directly model the data distribution.
Consequently, we choose to model the data distribution in the feature space, which is generally more manageable.
Our assumption is motivated by the Simplex-Encoded-Labels Interpolation (SELI)  \cite{Thrampoulidis2022} used to characterize the neural collapse phenomenon in imbalanced learning.
It can be inferred that the features tend to collapse towards the mean values of their corresponding classes.
Thus, we can assume that the feature norms of each class sample are equal and employ the von Mises-Fisher (vMF) distribution \cite{Mardia2000} on the unit sphere to represent the feature distribution.

\textbf{The vMF Distribution} is a fundamental probability distribution on the unit hyper-sphere $\mathbb{S}^{p-1}$ in $\mathbb{R}^p$.
Its probability density function for a random $p$-dimensional unit vector $\bm{z}$ is given by:
\begin{equation}
  \label{eq:vMF}
  f_{p}(\bm{z} |{\bm {\mu }},\kappa ) = \frac{1}{C_{p}(\kappa )}\exp \left({\kappa {\bm {\mu }}^T \bm {z} }\right), \quad C_{p}(\kappa ) = {\frac {(2\pi )^{p/2}I_{(p/2-1)}(\kappa )}{\kappa ^{p/2-1}}},
\end{equation}
where $\bm z$ is a $p$-dimensional unit vector, ${\kappa \geq 0,\left\Vert {\bm {\mu }}\right\Vert_2 =1}$ and 
$I_{(p/2-1)}$ denotes the modified Bessel function of the first kind at order $p/2-1$, which is defined as:
\begin{equation}
    \label{eq:Bessel}
    I_{(p/2-1)}(\kappa) = \sum_{i=0}^{\infty }{\frac{1}{i!\Gamma (p/2-1+i+1)}} (\frac{\kappa}{2})^{2i+p/2-1}.
\end{equation}
The parameters $\bm {\mu} $ and $\kappa$ are referred to as the mean direction and concentration parameters, respectively.
A higher concentration around the mean direction $\bm{\mu}$ is observed with greater $\kappa$, and the distribution becomes uniform on the sphere when $\kappa=0$.

Based on the above assumption and cross-entropy loss, we can obtain the optimization target and the Bayes classifier as:
\begin{equation}
    \label{eq:classifier}
      \mathcal{L}_{\name}(\bm{x}) = -\log p(y|\bm{z}), \quad p(y|\bm{z})=\frac{p(y)p(\bm{z}|y)}{\sum_{y'}p(y')p(\bm{z}|y')}=\frac{\pi_{y} \frac{1}{C_{p}(\kappa_y )}\exp \left({\kappa_y {\bm {\mu_y }}^T \bm {z} }\right)}{\sum_{y'}\pi_{y'} \frac{1}{C_{p}(\kappa_{y'})}\exp \left({\kappa_{y'} {\bm {\mu_{y'}}}^T \bm {z} }\right)},
\end{equation}

where $\bm{z}$ is the corresponding feature embedding of input $\bm{x}$, $\pi_y$ is the class frequency in the training or test set, $\kappa_y$ and $\bm {\mu_y}$ are the parameters of the vMF distribution for class $y$.
From \cref{eq:classifier}, it is evident that the classifier is a linear classifier within the feature space.
However, its fundamental distinction from existing methods lies in the explicit construction of the classifier based on the Bayes’ theorem and parameter estimation through maximum a posteriori estimation rather than implicit estimation through gradient descent.
Based on our distribution adjustment method, the empirical analysis is depicted in \cref{fig:kappa-norm} and \cref{tab:DA}.

\subsection{Maximum A Posteriori Estimation of the vMF Distribution}
\label{sec:MAP}

In the following section, we will present a method for estimating the parameters $\kappa_y$ and $\bm{\mu_y}$ in \cref{eq:classifier} using a point estimation approach during the training process.
Under the assumption of the vMF distribution, the parameters  can be estimated by maximum likelihood estimation (MLE).
However, during the early stages of training, the random distribution of features will lead to unstable optimization of the classifier.
To tackle this concern, we adopt the Maximum A Posteriori (MAP) estimation method to incorporate a prior distribution for parameter estimation.
This approach can be viewed as a regularized MLE.

\textbf{Conjugate Prior.} Suppose that a series of $N$ vectors $\{(\bm{z}_i)_i^N\}$ on the unit hyper-sphere $\mathbb{S}^{p-1}$ are independent and identically distributed (i.i.d.) observations from a vMF distribution.
The conjugate prior can be defined as:
\begin{equation}
    \label{eq:conjugate_prior}
    p(\bm{\mu}, \kappa) = \frac{1}{C} \frac{1}{C_p^{\alpha_0}(\kappa)} \exp(\beta_0 \kappa \bm{m_0}^T \bm{\mu}),
\end{equation}
where $\alpha_0 \ge 0, \beta_0 \ge 0, \bm{m_0} \in \mathbb{S}^{p-1}$ are the parameters of the prior distribution, and $C$ is an unknown normalization constant.

\textbf{Conjugate Posterior.} Given $ \bm{Z} = \{(\bm{z}_i)_i^N\}$, the posterior distribution of $\bm{\mu}$ and $\kappa$ takes the form:
\begin{equation}
    \label{eq:conjugate_posterior}
    p(\bm{\mu}, \kappa | \bm{Z}) = \frac{1}{C} \frac{1}{C_p^{\alpha}(\kappa)} \exp(\beta \kappa \bm{m}^T \bm{\mu}),
\end{equation}

where $\alpha = \alpha_0 + N, \beta = ||\beta_0 \bm{m_0} + \sum_{i=1}^N \bm{z}_i||_2$ and $\bm{m} = (\beta_0 \bm{m_0} + \sum_{i=1}^N \bm{z}_i)/{\beta}$.

\begin{proposition}[MAP Estimation]
    \label{prop:MAP}
    Suppose that a series of $N$ vectors $\{(\bm{z}_i)_i^N\}$ on the unit hyper-sphere $\mathbb{S}^{p-1}$ are independent and identically distributed (i.i.d.) observations from a vMF distribution. The maximum a posteriori (MAP) estimates of the mean direction ${\bm \mu}$ and concentration parameter $\kappa$ satisfy the following equations:
\begin{equation}
    \label{eq:map}
    \bm \mu = \bm{m}, \quad  \frac{I_{p/2}(\kappa)}{I_{p/2-1}(\kappa)} = \frac{\beta}{\alpha}.
\end{equation}
\end{proposition}

Derived from the MAP estimates of the vMF distribution, the parameters of conjugate prior can be interpreted in terms of pseudo-observations.
Specifically, $\bm{m_0}$ and $\beta_0$ represent the direction and length of the pseudo-observations, respectively.
Additionally, $\alpha_0$ denotes the number of pseudo-observations.
This can help choose reasonable hyperparameters for the prior distribution.

Based on the aforementioned MAP estimation, we can efficiently estimate the parameters of the Bayes classifier during the training process. A simple approximation \cite{Sra2012} to $\kappa$ is:
\begin{equation}
    \label{eq:mle_approx_kappa}
    {\hat {\kappa }}=\frac {p\beta \alpha}{\alpha^2-\beta^2}.
\end{equation}
Furthermore, the sample mean of each class is estimated in an online manner by aggregating statistics from the current mini-batch:
\begin{equation}
    \label{eq:online_mean}
\bar{\bm{{z}}}_j^{(t)} = \frac{n_j^{(t-1)}\bar{\bm{{z}}}_j^{(t-1)} + s_j^{(t)} \bar{\bm{z}}^{\prime(t)}_j} {n_j^{(t-1)} +s_j^{(t)}},
\end{equation}
where ${\bar{\bm{z}}_j}^{(t)}$ is the estimated sample mean of class $j$ at step $t$ and $\bar{\bm{z}}^{\prime(t)}_j$ is the sample mean of class $j$ in current mini-batch.
$n_j^{(t-1)}$ and $s_j^{(t)}$ are the sample numbers in the previous mini-batches and the current mini-batch, respectively.

\textbf{Prior Parameter.}
We now discuss the parameter settings for the prior distribution.
For $\bm{m_0}$,  we set $\bm{m_0}^y$ for each class $y$ to form a simplex equiangular tight frame (ETF) following ETF classifier \cite{Yang2022}. 
Therefore, we construct an ETF and obtain the respective $\bm{m_0}^y$, as shown below:
\begin{equation}
	\label{ETF_M}
	\mathbf{M}=\sqrt{\frac{K}{K-1}}\mathbf{U}\left(\mathbf{I}_K-\frac{1}{K}\mathbf{1}_K\mathbf{1}_K^T\right),
\end{equation}
where $\mathbf{M}=[\bm{m_0}^1,\cdots,\bm{m_0}^K]\in\mathbb{R}^{p\times K}$, feature dimension $p \ge (K-1)$, $\mathbf{U}\in\mathbb{R}^{p\times K}$ is a partial orthogonal matrix, $\mathbf{I}_K$ is the $K\times K$ identity matrix, and $\mathbf{1}_K$ is the $K$-dimensional vector of ones.
When $p < (K-1)$, $\mathbf{M}$ are calculated following \cite{Li2022a}.

Based on the preceding discussion, in order to ensure stability during the initial stages of training, we utilize gradient updates for the parameters $\bm{m_0}$ of the prior distribution.
It is worth noting that as training progresses, the influence of the prior distribution will gradually diminish.
Furthermore, as the strength of the prior distribution tends towards infinity, our approach will degenerate into a cosine classifier. The experimental results are reported in \cref{tab:DA}.

For parameters $\alpha_0$ and $\beta_0$, which represent the number and length of the pseudo-observations, respectively, a reasonable approach is to set them in proportion to the number of samples per class $N_y$.
Following this idea, we define new hyperparameters $\hat \alpha_0 = \alpha_0^y / N_y, \hat \beta_0 = \beta_0^y / N_y, y = 1, ..., K$.
We calculate all $\alpha_0^y$ and $\beta_0^y$ after selecting appropriate values for $\hat \alpha_0$ and $\hat \beta_0$.

\subsection{Overall Objective of \name}
\label{sec:method_detail}

As aforementioned in \cref{sec:MAP} , our method performs a computational estimation, which undergoes unstable optimization in the early training stages.
To this end, we integrate an LA classifier (a linear classifier trained with logit adjustment method) with a \name~classifier, both of which share a common backbone.
The introduction of the LA classifier facilitates stable training for \name.
In particular, we also employ an ensemble approach for prediction.
Furthermore, to reduce the coupling between the two classifiers during optimization, we employ a projection head specifically for the \name~classifier, and generate one view and two views of an input image for LA and \name~classifier respectively.
Finally, the loss functions are weighted and summed up as the overall loss function:
\begin{equation}
	\label{eq:loss}
	\mathcal{L} =  \mathcal{L}_{\text{\name}} +  \eta \mathcal{L}_{\text{LA}},
\end{equation}
where $\eta$ is the weight of the LA classifier. 

Importantly, $\mathcal{L}_{\text{LA}}$ can be replaced with any off-the-shelf long-tailed learning algorithms. In fact, we have observed that \name~is compatible with most existing methods, \emph{i.e.}, these methods can contribute to a stable convergence at earlier learning stages, while \name~considerably improves the final generalization performance. In other words, the gains of \name~ in terms of \emph{explicitly} modeling the Bayesian decision process are orthogonal to existing approaches, which are mainly designed under the principle of \emph{implicitly} estimating the Bayesian posterior probabilities. In this paper, we mainly report the improvements on top of LA due to its state-of-the-art performance.

\section{Experiment}
\label{sec:exp}
In this section, we conduct extensive experiments on multiple long-tail visual recognition benchmarks to validate the advantages of our method.

\subsection{Dataset and Evaluation Protocol}

We investigate the performance of our models on four prevalent long-tailed image classification datasets: CIFAR-10/100-LT, ImageNet-LT, and iNaturalist. We adopt the partition strategy proposed in \cite{Liu2022,Kang2020}, grouping the categories into three subsets based on the number of training samples: Many-shot categories with over $100$ images, Medium-shot categories with $20-100$ images, and Few-shot categories with fewer than $20$ images. For evaluation, the top-1 accuracy on the corresponding balanced validation or test sets is reported.

\textbf{CIFAR-10/100-LT.}\quad
CIFAR-10-LT and CIFAR-100-LT datasets are derived from their original counterparts, CIFAR-10 and CIFAR-100 \cite{krizhevsky2009learning}, by employing a sampling technique \cite{Cao2019,Cui2019}. Specifically, we adopt an exponential function $N_j\!=\!N\!\times\!\lambda^j$, where $\lambda\!\in\!(0,1)$. $N$ is the size of the original training set, and $N_j$ is the sample number in the $j$-th class. The balanced validation sets from the original datasets are used for testing. The imbalance degree in the datasets is measured by the imbalance factor $\gamma$, defined as $\gamma=\text{max}(N_j)/\text{min}(N_j)$. In our experiments, we set $\gamma$ to typical values of $10,50,100$.

\textbf{ImageNet-LT.}\quad
ImageNet-LT is a long-tailed version of the ImageNet dataset, constructed by sampling a subset of the original dataset following the Pareto distribution with power value $\alpha_p=6$ \cite{Liu2022}. The dataset contains $115.8$K images with $1,000$ classes, where each class has a varying number of images, ranging from $5$ to $1,280$. We used the standard setup for evaluation.

\textbf{iNaturalist 2018.}\quad
iNaturalist 2018~\cite{VanHorn2018} is a large-scale dataset that contains $437.5$K images from $8142$ different species. The dataset is highly imbalanced, with an imbalance factor $\gamma=500$. This makes iNaturalist an ideal dataset for evaluating long-tailed learning methods. We used iNaturalist to test the effectiveness of our method on real-world, complex datasets.

\subsection{Implementation Details}
\label{sec:implementation_details}
The training of all models involves the utilization of an SGD optimizer with a momentum of $0.9$.

\begin{table*}[t]
    \centering
    \tabcolsep=0.5cm
    \caption{Top-1 accuracy of ResNet-32 on CIFAR-100-LT and CIFAR-10-LT.
    $*$ denotes results borrowed from \cite{Zhou2020}.
    $\dagger$ denotes  our implementation.
    We report the results of 200 epochs.
    }
    \resizebox{0.85\textwidth}{!}{
    \begin{tabular}{l|ccc|ccc}
    \toprule[1pt]
     Dataset                           & \multicolumn{3}{c|}{CIFAR-100-LT} & \multicolumn{3}{c}{CIFAR-10-LT}\\
     \midrule
     Imbalance Factor                  & 100            & 50             & 10             & 100            & 50             & 10\\
     \midrule
     CB-Focal \cite{Cui2019}      & 39.6   & 45.2 & 58.0 & 74.6 & 79.3 & 87.5 \\
     LDAM-DRW$^*$ \cite{Cao2019} & 42.0   & 46.6 & 58.7 & 77.0 & 81.0 & 88.2 \\
     BBN \cite{Zhou2020}          & 42.6   & 47.0 & 59.1 & 79.8 & 81.2 & 88.3 \\
     SSP \cite{Yang2020}          & 43.4   & 47.1 & 58.9 & 77.8 & 82.1 & 88.5 \\
     VS \cite{kini2021label}            & 43.5 & - & - & 80.8     & -     & - \\
     TSC \cite{Li2022}            & 43.8   & 47.4 & 59.0 & 79.7 & 82.9 & 88.7 \\
     Casual model \cite{Tang2020} & 44.1   & 50.3 & 59.6 & 80.6 & 83.6 & 88.5 \\
     
     CDT \cite{ye2020identifying}           & 44.3  & - & 58.9 & 79.4     & -     & 89.4 \\
     ETF Classifier \cite{Yang2022}           & 45.3  & 50.4  & -  & 76.5     & 81.0     & - \\
     LADE \cite{hong2021disentangling}          & 45.4  & 50.5  & 61.7  & -     & -     & - \\
   
     MetaSAug-LDAM \cite{Li2021}  & 48.0   & 52.3 & 61.3 & 80.7 & 84.3 & 89.7 \\
     GCL \cite{MengkeLi2022}           & 48.7  & 53.6  & -  & 82.7     & 85.5     & - \\

     Logit Adj.$^\dagger$ \cite{Menon2021}  & 50.5 & 54.9 & 64.0 & 84.3 & 87.1 & 90.9 \\
     \midrule
     \cellcolor{lightgray!50}\name$^\dagger$             & \cellcolor{lightgray!50}\textbf{52.5} &\cellcolor{lightgray!50}\textbf{57.3} &\cellcolor{lightgray!50}\textbf{66.1} &\cellcolor{lightgray!50}\textbf{85.4} &\cellcolor{lightgray!50}\textbf{88.4} &\cellcolor{lightgray!50}\textbf{92.2} \\

    \bottomrule[1pt]
    \end{tabular}
    }
     \label{tab:cifar-lt}
\end{table*}

\begin{table}[t]
  \centering
  \begin{minipage}{0.5\textwidth}
    \centering
   \caption{Results on ImageNet-LT and iNaturalist 2018 (Res50/ResX50: ResNet-50/ResNeXt-50, 90-epoch training). $\dagger$ denotes our implementation.. }
    \resizebox{\textwidth}{!}{
 
    \begin{tabular}{l|cc|c}
    \toprule[1pt]
    \multirow{2}{*}{Method} &  \multicolumn{2}{c|}{ImageNet-LT} & iNaturalist 2018   \\
    \cmidrule(r){2-4}
                                          & Res50         & ResX50        & Res50\\
    \midrule
    
    $\tau$-norm\cite{Kang2020}      & 46.7          & 49.4          & 65.6\\
    MetaSAug \cite{Li2021}                 & 47.4          & --            & 68.8\\
    SSP~ \cite{Yang2020}                   & 51.3          & --            & 68.1\\
    ALA \cite{zhao2022adaptive}            & 52.4          & 53.3          & 70.7\\
    DisAlign \cite{Zhang2021}              & 52.9          & 53.4          & 69.5\\
    vMF classifier \cite{wang2022towards}  &  --           & 53.7          & --\\
    SSD \cite{Li2021a}                     & --            & 53.8          & 69.3\\
    ResLT \cite{Cui2022}                   & --            & 56.1          & 70.2\\
    Logit Adj.$^\dagger$ \cite{Menon2021}  & 55.1          & 56.5          & 71.0 \\
    \midrule
    \cellcolor{lightgray!50}\name$^\dagger$                       &\cellcolor{lightgray!50}\textbf{56.7} &\cellcolor{lightgray!50}\textbf{57.6} &\cellcolor{lightgray!50}\textbf{72.3}\\
    \bottomrule[1pt]
    \end{tabular}
    }
    \label{tab:imagenet}
  \end{minipage}
  \hspace{0.5ex}
  \begin{minipage}{0.48\textwidth}
    \centering
    \caption{Top-1 accuracy on ImageNet-LT. We report ResNet-50 results with 90-epoch and 180-epoch training.
        $\dagger$ denotes our implementation.}
         \resizebox{0.95\textwidth}{!}{
       \begin{tabular}{l|ccc|c}
        \toprule[1pt]
        Method                              & Many   & Medium & Few    & All \\
        \midrule
       \textit{90 epochs}\\
       $\tau$-norm \cite{Kang2020}           & 56.6   & 44.2   & 27.4   & 46.7\\
       DisAlign \cite{Zhang2021}             & 61.3   & 52.2   & 31.4   & 52.9\\
       DRO-LT \cite{Samuel2021}              & 64.0   & 49.8   & 33.1   & 53.5\\
       RIDE \cite{Wang2020}                  & 66.2   & 51.7   & 34.9   & 54.9\\
       Logit Adj.$^\dagger$ \cite{Menon2021} & 65.5   & 53.2   & 32.3   & 55.1\\
    
        \midrule
       \cellcolor{lightgray!50}\name$^\dagger$        &\cellcolor{lightgray!50}\textbf{66.3} &\cellcolor{lightgray!50}\textbf{54.3} &\cellcolor{lightgray!50}\textbf{38.1} &\cellcolor{lightgray!50}\textbf{56.7}\\
        \midrule
        
       \textit{180 epochs}\\
       Logit Adj.$^\dagger$ \cite{Menon2021}     & \textbf{68.0}    & 52.5          & 	34.2         & 56.0 \\
        \midrule
       \cellcolor{lightgray!50}\name$^\dagger$       &\cellcolor{lightgray!50}67.5 &\cellcolor{lightgray!50}\textbf{55.3} &\cellcolor{lightgray!50}\textbf{37.9} &\cellcolor{lightgray!50}\textbf{57.6} \\
       \bottomrule[1pt]
    \end{tabular}
    }
        \label{tab:imagenet-resnet50}
  \end{minipage}
\end{table}

\textbf{CIFAR-10/100-LT.}\quad
For long-tailed CIFAR-10 and CIFAR-100, we utilize ResNet-32 \cite{He2016} as the backbone network. For the \name~classifier, we employ a projection head with a hidden layer dimension of $512$ and an output dimension of $128$.
We apply AutoAug \cite{Cubuk2019} and Cutout \cite{DeVries2017} as data augmentation strategies for the LA classifier, while SimAug \cite{Ting2020} is utilized for the \name~classifier.
The loss weight is assigned equally ($\eta = 1$) to both classifiers.
We train the network for $200$ epochs with a batch size of $256$ and a weight decay of $4$e$-4$.
The prior parameters $\hat \alpha_0$ and $\hat \beta_0$ are set to $40$ and $8$, respectively.
We adopt a cosine schedule to regulate the learning rate. This approach entails gradually ramping up the learning rate to $0.3$ during the first 5 epochs, followed by a smooth factor applied that varies between $0$ and $1$ according to a cosine function.
Unless specified, our ablation study and analysis employ these training settings.
Additionally, we train the model for $400$ epochs with a similar learning rate schedule to enable a more thorough comparison.

\textbf{ImageNet-LT \& iNaturalist 2018.}\quad
We adopt ResNet-50 \cite{He2016} as the backbone network for both ImageNet-LT and iNaturalist 2018.
The \name~classifier is comprised of a projection head with an output dimension of $1024$ and a hidden layer dimension of $2048$,
while the LA classifier is employed as a cosine classifier \cite{Wang2018}. 
For data augmentation, we use RandAug \cite{Cubuk2020} and SimAug for the LA and \name~classifiers, respectively.
The prior parameters $\hat \alpha_0$ and $\hat \beta_0$ are set as 20 and 0.6 for ImageNet-LT, while $10$ and $0.3$ are set for iNaturalist 2018.
We also assign equal loss weight ($\eta = 1$).
The model is trained for $90$ epochs with a batch size of $256$ and a cosine learning rate schedule.
For ImageNet-LT, the initial learning rate is set to $0.1$ and the weight decay is set to $5$e$-4$.
We also train the model for $90$ epochs with ResNeXt-50-32x4d \cite{xie2017aggregated}, and for $180$ epochs with ResNet-50. 
For iNaturalist 2018, the initial learning rate is set to $0.2$ and the weight decay is set to $1$e$-4$.

\subsection{Main Results}

\begin{wraptable}{r}{0.56\textwidth}
    \centering
    \small
    \vskip -0.19in
    \caption{Top-1 accuracy of ResNet-32 on CIFAR-100-LT (imbalance factor: 100).
    $\dagger$ denotes our implementation.
    }
    \resizebox{1.0\linewidth}{!}{
   \begin{tabular}{l|ccc|c}
    \toprule[1pt]
     Method                             & Many   & Medium        & Few    & All  \\
    \midrule
    \textit{200 epochs} \\
    DRO-LT \cite{Samuel2021}              & 64.7   & 50.0          & 23.8   & 47.3\\
    RIDE \cite{Wang2020}                  & {68.1} & 49.2          & 23.9   & 48.0\\
    Logit Adj.$^\dagger$ \cite{Menon2021} & 67.2   & 51.9          & 29.5   & 50.5\\
    \midrule
    \cellcolor{lightgray!50}\name$^\dagger$                      &\cellcolor{lightgray!50}\textbf{68.7} &\cellcolor{lightgray!50}\textbf{53.2}  &\cellcolor{lightgray!50}\textbf{32.9} &\cellcolor{lightgray!50}\textbf{52.5}\\
    \midrule
    \textit{400 epochs} \\
    Logit Adj.$^\dagger$ \cite{Menon2021} & 68.1          & 53.0          & 32.4          & 52.1\\
    \midrule
    \cellcolor{lightgray!50}\name$^\dagger$                      &\cellcolor{lightgray!50}\textbf{70.2} &\cellcolor{lightgray!50}\textbf{55.0} &\cellcolor{lightgray!50}\textbf{34.1} &\cellcolor{lightgray!50}\textbf{54.1}\\
    \bottomrule[1pt]
\end{tabular}}
 \label{tab:cifar100-100}
\end{wraptable}
\textbf{CIFAR-10/100-LT.}\quad
The comparison between \name~and existing methods on CIFAR-100-LT and CIFAR-10-LT are summarized in \cref{tab:cifar-lt}. Our \name~significantly outperforms the competitors, demonstrating its efficacy in long-tailed classification. 
We also present extended and comprehensive results in \cref{tab:cifar100-100}, ensuring the preservation of an imbalance factor of 100. 
Specifically,  compared to Logit Adj., our approach demonstrates a notable enhancement of $3.4\%$ for the tail classes with 200 epochs.
Furthermore, our method achieves a consistent improvement of approximately $2.0\%$ across all subsets with 400 epochs.

\textbf{ImageNet-LT.}\quad
We present comprehensive evaluation results of our approach on ImageNet-LT in \cref{tab:imagenet}. Leveraging the ResNet-50 and ResNeXt-50 backbones with $90$ epochs training,  \name~ surpass Logit Adj. by a margin of $1.6\%$ and $1.1\%$ respectively. 
Furthermore, \cref{tab:imagenet-resnet50} lists detailed results on more training
settings for ImageNet-LT dataset.
Notably, \name~significantly outperforms  the Logit Adj. in tail classes, exhibiting improvements of $5.7\%$. 
With 180 epochs, \name~exhibits a marginal decline of $0.5\%$ in Top-1 accuracy for head classes, whereas it demonstrates an improvement of $3.7\%$ for tail classes, resulting in an overall accuracy enhancement of $1.6\%$.

\textbf{iNaturalist 2018.}\quad
\cref{tab:imagenet} also presents the experimental results obtained from implementing our \name~on the iNaturalist 2018 dataset. Due to its highly imbalanced nature, iNaturalist 2018 serves as an exemplary platform to investigate the influence of imbalanced datasets on the performance of machine learning models. Our \name~outperforms Logit Adj. by $1.3\%$ under the same setting.

\subsection{Effectiveness of \name}
\label{sec:effectiveness}

\begin{table}[t]
  \caption{Analysis of distribution adjustment. \cmark denotes we perform adjustment during this stage.}
   \centering
   \resizebox{0.7\textwidth}{!}{
\begin{tabular}{l|cc|ccc|c}
   \toprule[1pt]
Adjusted Classifier   & Training & Testing & Many & Medium & Few & All \\
    \midrule
      w/o Adjustment   &          &         & 69.5     & 51.9       &  28.0    &  50.9   \\
    \midrule
\multirow{2}{*}{LA \cite{Menon2021}}   & \cmark         &  \cmark       & 66.2 & 51.1   & 28.9 & 49.7  \\
                      &          & \cmark        &   67.5   &   48.6     &  25.7    &  48.3   \\
      \midrule
\multirow{2}{*}{BAPE}  &  \cmark        &  \cmark      & 68.9 & 53.6 & 28.5 & 51.3 \\
                        &   \cellcolor{lightgray!50}       &   \cellcolor{lightgray!50}\cmark      & \cellcolor{lightgray!50}68.7 & \cellcolor{lightgray!50}53.2 & \cellcolor{lightgray!50}\textbf{32.9} & \cellcolor{lightgray!50}\textbf{52.5}\\                 
     \bottomrule[1pt]
\end{tabular}
}
\label{tab:DA}
\end{table}

\begin{wraptable}{r}{0.56\textwidth}
    \centering
    \small
    \vskip -0.15in
    \caption{Ablation study on CIFAR-100-LT. DA denotes the Distribution Adjustment.}
    \resizebox{1.0\linewidth}{!}{
   \begin{tabular}{l|ccc|c}
       \toprule
Ablation              & Many & Medium & Few & All \\
  \midrule
w/o Prior \& DA & 66.1 & 48.2   & 24.2  &   47.3    \\
w/o Prior        & 64.2 & 49.9   & 27.2  &  48.1     \\
w/o DA           & \textbf{69.5} & {51.9}   & {28.0}  &   {50.9}    \\
\cellcolor{lightgray!50}Ours              &\cellcolor{lightgray!50}{68.7}& \cellcolor{lightgray!50}\textbf{53.2}   & \cellcolor{lightgray!50}\textbf{32.9}  &  \cellcolor{lightgray!50}\textbf{52.5}    \\
         \bottomrule           
\end{tabular}
}
\label{tab:ablation}
\end{wraptable}

\textbf{Ablation Study.}
In this section, we conduct experiments to validate different design choices of \name~as reported in \cref{tab:ablation}.
We first implement an initial version of \name~without incorporating prior samples or distribution adjustment (DA), which yields a result of $47.3\%$.
Subsequently, DA is applied to address the discrepancies in conditional distributions between the training and test sets. 
The adjusted $\kappa$  leads to a performance increase of $0.8\%$. 
We then investigate the effectiveness of prior distribution. As expounded in \cref{sec:MAP}, we generate a group of prior parameters, which guarantees more stable training, resulting in a significant enhancement of $3.6\%$.
Moreover, when combined with each other, DA and Prior exhibit substantial improvements in performance, with a respective increase of $1.6\%$ and $4.4\%$.
This finding demonstrates the seamless integration and synergistic effect of these two components, resulting in a further enhancement of overall performance.

\textbf{Analysis of Distribution Adjustment.}
In this section, we embark on an in-depth exploration of the Distribution Adjustment (DA) element of our framework.
The DA operates in a similar vein to the idea of post-hoc normalization on both weight and feature, albeit with a distinction that we explicitly model the norms as a parameter $\kappa$, which signifies the degree of concentration in the vMF distribution.
In light of this observation, we conduct a series of experiments on multiple variants of \name~reported in \cref{tab:DA}, employing diverse adjustment (normalization on both weight and feature for LA) configurations within the LA classifier or the \name~classifier.
All results are on CIFAR-100-LT.

\begin{figure}[t]
	\centering
 	\begin{minipage}{0.38\linewidth}
		\centering
		\includegraphics[width=1.0\linewidth]{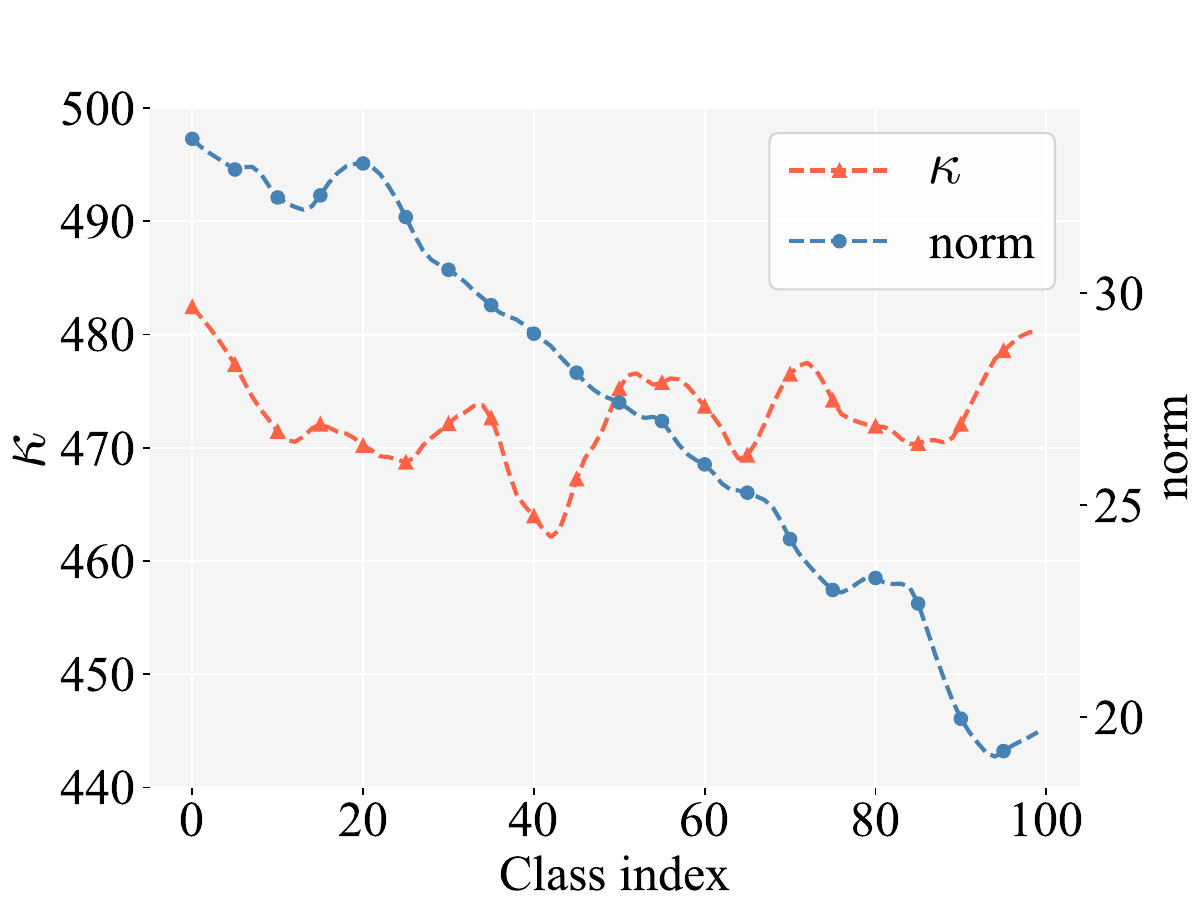}
		\caption{$\kappa$ (\name) and the norm (products of weight norm and feature norm in LA) of different classes.}
        \label{fig:kappa-norm}
	\end{minipage}
 \hfill
    \begin{minipage}{0.6\linewidth}
        \centering
        \includegraphics[width=1.0\linewidth]{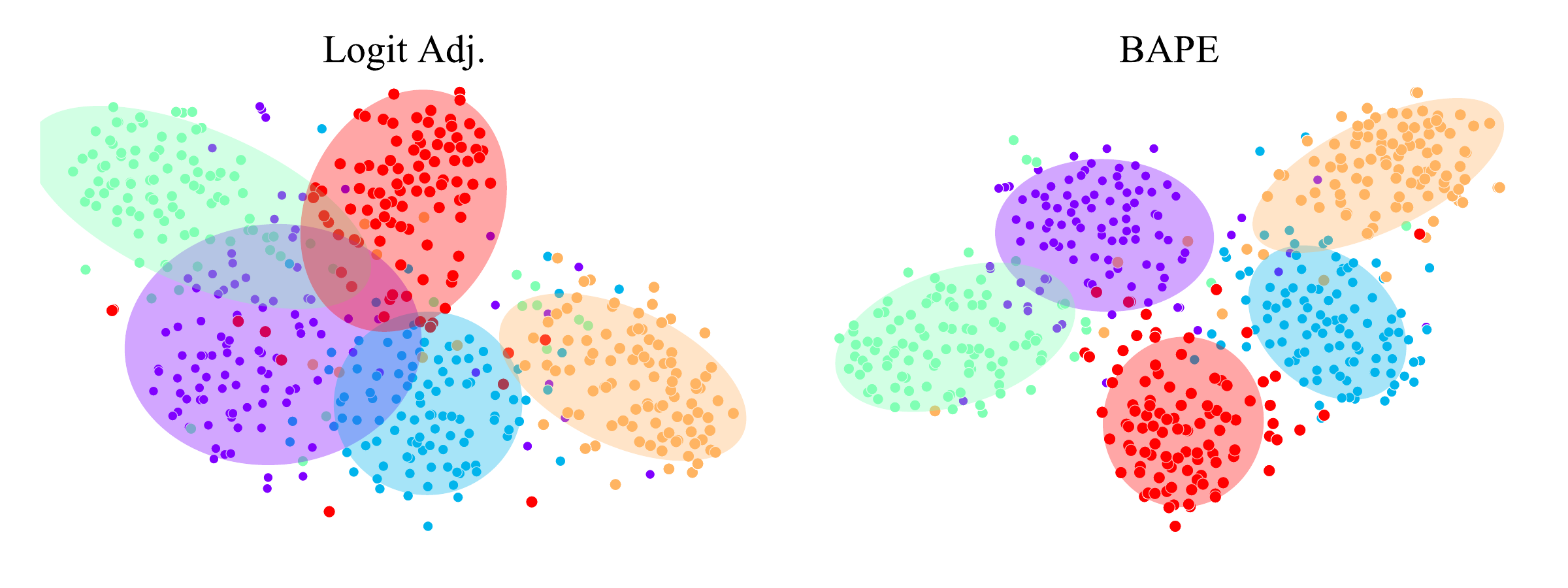}
        \caption{Visualization of the feature embedding via t-SNE. We take the mean of each category as the center and calculate the axis length and angle by utilizing the eigenvalues and eigenvectors obtained from the feature covariance matrix. This enables us to draw an ellipse capable of encompassing the majority of feature points within each category.}
        \label{fig:t-SNE}
	\end{minipage}
\end{figure}

To evaluate the performance of different settings, we employ the unadjusted version of \name~as the reference standard. 
It is worth noting that applying adjustment to the LA classifier results in a weakened behavior, while employing adjustment to the \name~classifier yields an improvement of accuracy.
This observation provides valuable insight into the superiority of our method, which \emph{explicitly} estimates the classifier parameters.
For a clear illustration, \cref{fig:kappa-norm} presents a visualized comparison between the norms (product of weight norm and feature norm) computed from the LA classifier and the $\kappa$ parameters within the \name~classifier of an optimized model.
We can observe that the LA classifier is biased and yields norms strongly correlated with class frequency, while the \name~classifier effectively overcomes that imbalance and focuses on learning the essential, rather than the frequency of each class, therefore learning frequency independent $\kappa$.

We also apply adjustment to LA during both stages and adopt an appropriate temperature parameter,
which is equivalent to employing a cosine classifier \cite{Wang2018}.
Moreover, performing adjustment to \name~during both stages leads to a slight decrease in accuracy, especially for tail classes.
This emphasizes the necessity to preserve $\kappa$ during the training stage.

\textbf{Visualization of Feature Embeddings.}
Our approach enables the explicit estimation of the parameters of the Bayesian classifier without employing gradient descent.
As a result, it effectively mitigates the issue of imbalance gradient.
To depict this, we illustrate a t-SNE \cite{van2008visualizing} visualization of the feature embeddings of five tail classes for optimized Logit Adj. and \name~in \cref{fig:t-SNE}.
It can be observed that \name~achieves a more distinct separation of different tail features compared to the Logit Adj., enabling the classifier to discern them correctly.


\section{Conclusion}
\label{sec:conclusion}

In this study, we introduce an innovative methodology, termed BAPE (Bayes Classifier by Point Estimation), which is devised for the \emph{explicit} learning of a Bayes classifier. 
In contrast to prevailing implicit estimation methods, BAPE ensures a superior theoretical approximation of the data distribution by explicitly modeling the parameters and employing point estimation.
Notably, BAPE offers dual benefits: it learns the Bayes classifier directly using Bayes' theorem and it circumvents the need for gradient descent. 
This two-pronged strategy guarantees an optimal decision rule while also addressing the issue of gradient imbalance.
Furthermore, we introduce a straightforward yet potent method to adjust the distribution.
This method enables our Bayes classifier, derived from an imbalanced training set, to adapt effectively to a test set with an arbitrary imbalance factor.
Importantly, this adjustment technique enhances performance without accruing additional computational costs. 
Furthermore, we demonstrate that the advantages of our approach are independent of existing learning methodologies tailored for long-tailed scenarios, as the majority of these approaches are primarily constructed based on the principle of implicitly estimating posterior probabilities.
Comprehensive experiments were conducted on popular benchmarks including CIFAR-LT-10/100, ImageNet-LT, and iNaturalist2018. The results offer strong evidence of the effectiveness and superiority of BAPE.


\bibliographystyle{plain}
\bibliography{MyBib}

\end{document}